\documentclass[conference]{IEEEtran}
\IEEEoverridecommandlockouts
\usepackage{cite}
\usepackage{amsmath,amssymb,amsfonts}
\usepackage{algorithmic}
\usepackage{algorithm}
\usepackage{graphicx}
\newtheorem{definition}{Definition}
\usepackage{algorithmic}
\newtheorem{problem}{Problem}
\usepackage{textcomp}
\usepackage{hyperref}
\usepackage{xcolor}
\def\BibTeX{{\rm B\kern-.05em{\sc i\kern-.025em b}\kern-.08em
    T\kern-.1667em\lower.7ex\hbox{E}\kern-.125emX}}
\begin{document}

\title{Stable and Safe Human-aligned Reinforcement Learning through Neural Ordinary Differential Equations
}

\author{\IEEEauthorblockN{Liqun Zhao}
\IEEEauthorblockA{\textit{Department of Engineering Science} \\
\textit{University of Oxford}\\
Oxford, UK \\
liqun.zhao@eng.ox.ac.uk}
\and
\IEEEauthorblockN{Keyan Miao}
\IEEEauthorblockA{\textit{Department of Engineering Science} \\
\textit{University of Oxford}\\
Oxford, UK \\
keyan.miao@eng.ox.ac.uk}
\and
\IEEEauthorblockN{Konstantinos Gatsis}
\IEEEauthorblockA{\textit{School of Electronics and Computer Science} \\
\textit{University of Southampton}\\
Southampton, UK \\
k.gatsis@soton.ac.uk}
\and
\IEEEauthorblockN{Antonis Papachristodoulou}
\IEEEauthorblockA{\textit{Department of Engineering Science} \\
\textit{University of Oxford}\\
Oxford, UK \\
antonis@eng.ox.ac.uk}
}

\maketitle

\begin{abstract}
Reinforcement learning (RL) excels in applications such as video games, but ensuring safety as well as the ability to achieve the specified goals remains challenging when using RL for real-world problems, such as human-aligned tasks where human safety is paramount. This paper provides safety and stability definitions for such human-aligned tasks, and then proposes an algorithm that leverages neural ordinary differential equations (NODEs) to predict human and robot movements and integrates the control barrier function (CBF) and control Lyapunov function (CLF) with the actor-critic method to help to maintain the safety and stability for human-aligned tasks. Simulation results show that the algorithm helps the controlled robot to reach the desired goal state with fewer safety violations and better sample efficiency compared to other methods in a human-aligned task.
\end{abstract}


\section{Introduction}
Safety in robotics has become a focal point in current research activity \cite{wang2023trustworthy,cao2023physical,wang2023quadue,he2022robust}, and RL approaches have been applied to ensure safety in human-robot scenarios \cite{zacharaki2020safety,liu2021deep}. Recently, an increasing interest has emerged in integrating control approaches with RL to help ensure safety in real-world tasks, and concepts like control barrier function (CBF) \cite{cheng2019end,wang2022safety,do2023game} have been used as safety constraints. However, in many studies, the ability of the robot to achieve the designed goal state is not considered, and a physics-based control-affine nominal model of the system is required \cite{cheng2019end,emam2019robust}.\\
Maintaining stability for a control system is also paramount in human-aligned tasks, and concepts like Lyapunov functions are currently used in learning to help guarantee the stability \cite{han2020actor,cao2023physical2,miao2023towards,wang2023model}. However, these model-free algorithms are often data-intensive, and therefore, an algorithm with a higher sample efficiency may be desired for better applications in real-world scenarios when human exists.\\
\textbf{Our contributions.} 1. We introduce a primary controller that combines the CBF and Control Lyapunov Function (CLF) with the Soft Actor-Critic (SAC) algorithm \cite{haarnoja2018soft2} for human-aligned tasks where human and robot movements are approximated by neural ordinary differential equations (NODEs) \cite{chen2018neural}. 2. We propose an RL-based backup controller that prioritizes safety constraints when satisfying both safety and stability constraints simultaneously is not possible. 3. We combine the two controllers together to propose a new algorithm and show its better performance on one simulation scenario where human drivers exist compared to other baselines.

\section{Background}
\label{sec:Background}
\subsection{Preliminaries}
\subsubsection{Markov Decision Process} 
A Markov decision process (MDP) is defined by the tuple $\mathcal{M}$ which is $(\mathcal{X} , \mathcal{U},\mathcal{F}, r, c,\gamma, \gamma_{c} )$. $\mathcal{X} \subset \mathbb{R} ^{n} $ and $\mathcal{U} \subset \mathbb{R} ^{m}$ are state and control signal spaces, and $x_{t_k} \in \mathcal{X} $ is the state at timestep $t_k$, $u_{t_k} \in \mathcal{U} $ is the control signal at timestep $t_k$. $\mathcal{F}$ denotes the dynamics of the whole system including both human and the controlled robot, and by applying the control signal $u_{t_k}$ between timesteps $t_k$ and $t_{k+1}$, there is:
\begin{align}\label{expression of next state}
            x_{t_{k+1}} = x_{t_k} +  \int_{t_k}^{t_{k+1}} \mathcal{F}(\chi,x_{\chi},u_{t_k})  d\chi .
            \end{align}
The reward and cost are denoted as $r$ and $c$, and $\gamma$ and $\gamma_{c}$ are the discount factors. The transition probability is defined as $P(x_{t_{k+1}}|x_{t_k},u_{t_k})\triangleq I_{\{x_{t_{k+1}} = x_{t_k} +  \int_{t_k}^{t_{k+1}} \mathcal{F}(\chi,x_{\chi},u_{t_k})  d\chi  \}}$, where $I$ is a function that equals 1 if $x_{t_{k+1}}$ satisfies Eq.(\ref{expression of next state}), and 0 otherwise. Following Han et al. \cite{han2020actor}, the closed-loop transition probability is denoted as $P_{\pi}(x_{t_{k+1}}|x_{t_k})\triangleq \int_{\mathcal{U} }\pi(u_{t_k}|x_{t_k})P(x_{t_{k+1}}|x_{t_k},u_{t_k})du_{t_k}$. Additionally, the closed-loop state distribution at timestep $t_k$ is denoted by $\upsilon(x_{t_k}|\rho,\pi,t_k)$, which is calculated as $\upsilon(x_{t_{k+1}}|\rho,\pi,t_{k+1})=\int_{\mathcal{X}}P_{\pi}(x_{t_{k+1}}|x_{t_k})\upsilon(x_{t_k}|\rho,\pi,t_k)dx_{t_k}, \,\forall t_k \in \mathbb{N}$. The initial state distribution is $\upsilon(x_{t_0}|\rho,\pi,t_0)=\rho$.

\subsubsection{Definitions of Safety and Stability}
If there are $m$ safety constraints (for example, the controlled robot is required to avoid colliding with $m$ human users), the system is safe if $h_i(x_{t_k})  \geq 0$, $\forall t_k \geq t_0$ holds for $i = 1,\ldots, m$. Here $h_i: \mathbb{R}^n \rightarrow \mathbb{R} $ is a function corresponding to the $i^{th}$ safety constraint, and a safe set $\mathcal{C}_{i,0} \subset \mathbb{R} ^{n}$ is defined as follows:
\begin{equation}\label{ith safe set}
        \mathcal{C}_{i,0} = \{x_{t_k} \in \mathcal{X}|h_i(x_{t_k}) \geq 0\}.
\end{equation}
We require the system state to remain within this set $\mathcal{C}_{i,0}$. When the relative degree of the constraint $h_i(x_{t_k})   \geq 0$ is $r$, the discrete-time CBF can be used to ensure the forward invariance of the safe set, and a list of functions can be defined:
\begin{align}\label{list of functions for relative degree}
  \Phi_{i,0}(x_{t_k}) & := h_{i}(x_{t_k}) \nonumber \\
  \Phi_{i,1}(x_{t_k}) & := \Delta \Phi_{i,0}(x_{t_k},u_{t_k}) + \kappa_{i,1}(\Phi_{i,0}(x_{t_k}))  \nonumber \\
  &\,\,\,\vdots  \nonumber \\
  \Phi_{i,r}(x_{t_k}) & := \Delta \Phi_{i,r-1}(x_{t_k},u_{t_k}) + \kappa_{i,r}(\Phi_{i,r-1}(x_{t_k}))
\end{align}
where $\Delta \Phi_{i,j}(x_{t_k},u_{t_k}):= \Phi_{i,j}(x_{t_{k+1}}) - \Phi_{i,j}(x_{t_k})$, $j=0,1,\dots,r\!-\!1$, and $\kappa_{i,j}(\cdot)$ are class $\mathcal{K}$ functions. Then the definition of the discrete-time CBF can be given as:
\begin{definition}[Discrete-time Control Barrier Function \cite{xiong2022discrete}] 
        For the system described by Eq.(\ref{expression of next state}), the function $h_i: \mathbb{R}^n \rightarrow \mathbb{R}$ is called a discrete-time CBF of relative degree $r$ if there exists $\Phi_{i,j}(x_{t_k})$, $j=0,1,\dots,r$ defined by Eqs.(\ref{list of functions for relative degree}) and $\mathcal{C}_{i,j}$ which are defined similarly to Eq.(\ref{ith safe set}), $j=0,1,\dots,r-1$ such that for all $x_{t_k}\in \bigcap_{j=0}^{r-1}\mathcal{C}_{i,j}$,
        \begin{equation}\label{CBF constraints}
                \Phi_{i,r}(x_{t_k}) \geq 0.
        \end{equation} 
\end{definition}
A controller satisfying Ineq.(\ref{CBF constraints}) can make the set $\bigcap_{j=0}^{r-1}\mathcal{C}_{i,j}$ forward invariant, and therefore, safety is maintained if there exists a controller such that $\forall i \in [1,m]$, Ineq.(\ref{CBF constraints}) holds for all $x_{t_k}\in \bigcap_{i=1}^{m} \bigcap_{j=0}^{r-1}\mathcal{C}_{i,j}$.
\\
In a stabilization task,  the system state is required to finally reach a goal state (for example. keeping a specified distance from a human user). Therefore, we define the cost signal as $c(x_{t_k},u_{t_k})=\left\lVert x_{t_{k+1}}-x_{\text{desired}}\right\rVert$ where $x_{t_{k+1}}$ is the next state following Eq.(\ref{expression of next state}), and $x_{\text{desired}}$ denotes the goal state (for example, the desired distance). Furthermore, define the cost function under the controller $\pi$ as $c_{\pi}(x_{t_k}) =  \mathbb{E}_{u_{t_k} \sim \pi}c(x_{t_k},u_{t_k})=
        \mathbb{E}_{u_{t_k} \sim \pi}[\left\lVert x_{t_{k+1}}-x_{\text{desired}}\right\rVert ]$,
and naturally, we expect that the value of $c_{\pi}(x_{t_k})$ decreases as $t_k$ increases, and achieve $c_{\pi}(x_{t_k}) = 0$ eventually, which means the agent reaches the goal state. Similar to Han et al. \cite{han2020actor}, we provide the following definition.

\begin{definition}[Stability in Mean Cost]\label{Stability in Mean Cost at the Equilibrium}
Let $\upsilon$ denote the closed-loop
state distribution, the goal state is said to be stable in mean cost if there exists a positive constant $b$ such that the condition $\lim_{t_k\to\infty}\mathbb{E}_{x_{t_k} \sim \upsilon}\big[c_{\pi}(x_{t_k})\big]=0$ holds for any initial state $x_{t_0} \in \{x_{t_0}|c_{\pi}(x_{t_0}) < b\}$. If $b$ is arbitrarily large,
the goal state is globally stable in mean cost.
\end{definition}

\subsubsection{Introduction to Neural Ordinary Equations (NODEs)}
According to Eqs.(\ref{list of functions for relative degree}), applying CBFs requires the system dynamics to obtain the future states $\{x_{t_{k+1}},x_{t_{k+2}},\dots, x_{t_{k+r}}\}$ of the whole system including both human and the controlled robot. However, in many cases, obtaining system dynamics or even a nominal model directly based on the physics law is difficult. Neural networks are important tools to estimate the dynamics, and there has been a rise in considering neural network hidden layers as states \cite{observer}. Chen et al. \cite{chen2018neural} introduces an ODE to approximate dynamics by neural networks as follows:
\begin{equation}
\left\{
\begin{array}{ll}
\dot {\hat {x}}_{t_k} = \mathcal{F}_\psi \left(t_k, {\hat {x}}_{t_k}, u_{t_k}\right) & \\
{\hat {x}}_{t_0} = x_{t_0}
\end{array}
\right.
\label{node}
\end{equation}
where $\mathcal{F}_\psi$ is a neural network with parameter $\psi$. By assuming a constant control signal, i.e., $u_\chi = u_{t_k}, \chi \in [t_k, t_{k+1})$, the inference of NODEs is:
\begin{equation}
    \hat x_{t_{k+1}} = \hat x_{t_k} +\int_{{t_k}}^{t_{k+1}} \mathcal{F}_\psi \left(\chi, \hat x_\chi, u_{t_k} \right)d\chi 
    \label{approximation of next state}
\end{equation}
which is used to estimate Eq.\eqref{expression of next state}. Then we can approximate all future states required for constructing Ineq.(\ref{CBF constraints}) by iteratively using Eq.(\ref{approximation of next state}).\\

\subsection{Definition of the Safe and Stable Control Problem}
Similar to Dawson et al. \cite{dawson2022safe1}, here we define:
\begin{problem}[Safe and Stable Control Problem]\label{Safe and Stable Control Problem}
For system described by Eq.(\ref{expression of next state}), given a desired goal state $x_{\text{desired}}$, a set $\mathcal{X}_0$  which is the set of  $x_{t_0}$ denoting the initial states, an unsafe set $\mathcal{X}_{\text{unsafe}}\subseteq \mathcal{X}$, and a safe set $\mathcal{X}_{\text{safe}}\subseteq \mathcal{X}$ such that $x_{\text{desired}} \in \mathcal{X}_{\text{safe}}$ and $\mathcal{X}_{\text{safe}}\cap\mathcal{X}_0 \neq \emptyset$, find a controller $\pi$ producing sequence $\{u_{t_k}\}_{{t_k} \geq t_0}$ such that the state sequence $\{x_{t_k}\}_{{t_k} \geq t_0}$ satisfying Eq.(\ref{expression of next state}) and $x_{t_0} \in \mathcal{X}_{\text{safe}}\cap\mathcal{X}_0$ satisfy: 1. \textbf{Safety}: $x_{t_k} \in \mathcal{X}_{\text{safe}}\,\,\,\,\forall t_k \geq t_0 $. 2. \textbf{Stability in Mean Cost at the Equilibrium:} $\lim_{t_k\to\infty}\mathbb{E}_{x_{t_k} \sim \upsilon}\big[c_{\pi}(x_{t_k})\big]=0$.

\end{problem}
When an exact expression of the real dynamics based on physics law is not available, we can utilize NODEs to approximate the real dynamics to construct constraints. See Section~\ref{sec:framework design} for a detailed description.

\section{Framework Design}
\label{sec:framework design}
\subsection{Learning the Dynamics via NODEs and Value Function of the Cost}
A good approximated system dynamics to predict the human and controlled robot movements is beneficial to the overall training. To obtain the NODE model, we collect state sequences $X =\{x_{t_k}, x_{t_{k+1}},\ldots, x_{t_{k+h}}\}$ and control sequences $U = \{u_{t_k}, u_{t_{k+1}},\ldots, u_{t_{k+h-1}}\}$ during the real interaction process following the real dynamics. The model then computes the approximated state sequence based on Eq.\eqref{approximation of next state}. Model loss is designed as $\ell = \frac{1}{h}\sum_{i=1}^h \lvert x_{t_{k+i}} - \hat x_{t_{k+i}}\rvert$, and the training process is illustrated as Algorithm \ref{alg:node-sysid}.
\\To help maintain stability for the system, inspired by commonly-used value functions in RL, we define the value function of the cost at the state $x_{t_k}$ as $L_{\pi}(x_{t_k})=\mathbb{E}_{\tau \thicksim \pi}\big[\sum_{i=0}^{\infty}\gamma_c^{i}c_{\pi}(x_{t_{k+i}})\big]$, where $\tau$ is a trajectory under controller $\pi$ starting from the initial state $x_{t_k}$. Based on this definition, $L_{\pi}(x_{t_k})$ can also be approximated by a neural network, and a natural strategy to achieve stability is to introduce a condition that drives the value of $L_{\pi}(x_{t_k})$ to decrease along the trajectory $\tau$. Inspired by the concept of CLF, the following condition proposed in Zhao et al. \cite{zhao2023barrier} can be utilized:
        \begin{align}\label{stability requirement 2}
                \mathbb{E}_{x_{t_k}\!\sim \mu_{\pi},x_{t_{k+1}}\!\sim \!P_{\pi}}\big[\!L_{\pi}(x_{t_{k+1}}) \! - \! L_{\pi}(x_{t_k})\!\big]
                \! \! \leq \! \!-\beta\mathbb{E}_{x_{t_k}\!\sim \!\mu_{\pi}}\big[\!L_{\pi}(x_{t_k})\!\big].
        \end{align}
$\mu_{\pi}(x)  \triangleq \lim_{N\to\infty}\frac{1}{N}\sum_{k=0}^{N}\upsilon(x_{t_k}=x|\rho,\pi,t_k)$ is the sampling distribution. Later, we introduce a method to obtain a controller leading to an $L_{\pi}$ satisfying Ineq.(\ref{stability requirement 2}).

\begin{algorithm}
    \caption{NODE-based Dynamics Learning for Human and Controlled Robot Movements Prediction}\label{alg:node-sysid}
    \begin{algorithmic}[1]
     \renewcommand{\algorithmicrequire}{\textbf{Input:}}
     \renewcommand{\algorithmicensure}{\textbf{Output:}}
    \REQUIRE  Learning rate $\eta_{1}$
    \STATE Collect trajectories $X$ and control variable sequences $U$
    \STATE Initialize dynamics model $\mathcal{F}_{\psi}$.
    \FOR{$i=1$ \textbf{to} $h$}
        \STATE $\hat x_{t_{k+i}} = \hat x_{t_{k+i-1}} +\int_{t_{k+i-1}}^{t_{k+i}} \mathcal{F}_\psi \left(\chi, \hat x_\chi, u_{t_{k+i-1}} \right)d\chi$
    \ENDFOR
    \STATE $\psi \leftarrow \psi - \eta_{1} \nabla_{\psi} \ell$
    \ENSURE  $\psi$
    \end{algorithmic} 
\end{algorithm}

\subsection{Augmented Lagrangian Method for Parameter Updating}
We denote the parameters of the RL-based primary controller and two action-value networks $Q^{\pi_p}$ by $\theta_{p}$ and $\phi_{i}$, where $i=1,2$, respectively. Additionally, we employ $L_{\nu}$, referred to as the Lyapunov network, to approximate $L_{\pi_{p}}$ with parameters $\nu$. According to the previous sections, by calculating expected values with predicted states, the RL problem with CBF and CLF constraints can be formed as follows:
\begin{equation}\label{RL-based constrained optimization with inequality constraints}
        \begin{split}
            \min_{\theta_{p}} \,\, & -V^{\theta_{p}}\\
            s.t. 
            \,\,& \mathbb{E}_{x_{t_k}\sim \mu_{\pi_{p}},u_{t_k}\sim \pi_p} \! \big[ -\Phi_{i,r}(x_{t_k})\big]  \! \leq  \! 0 \qquad \forall i \in [1,m],\vspace{1ex} \\\vspace{1ex}
            \,\,& \mathbb{E}_{x_{t_k}\sim \mu_{\pi_{p}},u_{t_k}\sim \pi_p} \! \big[ L_{\nu}(\hat{x}_{t_{k+1}})  \! -  \! L_{\nu}(x_{t_k})  \! +  \! \beta L_{\nu}(x_{t_k})\big]  \! \leq  \! 0, 
        \end{split}
    \end{equation}
where  $\mu_{\pi_{p}}(x)$ is the sampling distribution under the controller $\pi_{p}$, $-V^{\theta_{p}}$ is the objective function commonly used in SAC. Loss functions for updating the action-value networks $Q^{\pi_p}_{\phi_{i}}, i=1,2$, coefficient $\alpha_{p}$, and Lyapunov network $L_{\nu}$ are:
        \begin{align}
                \!\!\!\!&\!J_{Q^{\pi_p}}(Q^{\pi_p}_{\phi_i})\!\!=\!\mathbb{E}_{(x_{t_k}, u_{t_k}, r_{t_k}, x_{t_{k+1}})\sim \mathcal{D},\xi \sim \mathcal{N}}\!\!\Big[\!\!\big[r_{t_k} \!\!+\!\! \gamma \!\big(\!\!\mathop {\min }\limits_{j=1,2}\!\!Q^{\pi_p}_{targ,\phi_j}\!(x_{t_{k+1}}\!, \nonumber\\
                \!\!\!\!&\tilde{u}_{\theta_{p}}(x_{t_{k+1}},\xi ))\!-\!\alpha_{p}\! \log\! \pi_{\theta_{p}}\!(\tilde{u}_{\theta_{p}}\!(x_{t_{k+1}},\xi )|x_{t_{k+1}}\!)\big)\!-\!Q_{\phi_{i}}^{\pi_p}(x_{t_k},u_{t_k})\big]^2\!\Big], \label{RL-based controller value function loss}\\
                \!\!\!\!&\!J_{\alpha_{p}}(\alpha_{p})\!\!=\! - \alpha_{p} \times \mathbb{E}_{x_{t_k}\sim \mu_{\pi_{p}},\xi \sim \mathcal{N}}\!\big[\log \pi_{\theta_{p}}(\tilde{u}_{\theta_{p}}(x_{t_k},\xi )|x_{t_k}) \!+\!\mathcal{H}\big], \label{RL-based controller alpha function loss}\\
                \!\!\!\!&\!J_{L}(L_{\nu})\!\!=\!\mathbb{E}_{(x_{t_k}, c_{t_k}, x_{t_{k+1}})\sim \mathcal{D}}\!\Big[\!\big[c_{t_k} \!+\! \gamma_c L_{targ,\nu}(x_{t_{k+1}}) 
                \!-\!L_{\nu}(x_{t_k})\big]^2\!\Big],\label{Lyapunov  function loss}
        \end{align}
where $Q_{targ,\phi_i}, i=1,2$ are the target action-value networks, $\mathcal{H}$ is a predefined threshold, and $\mathcal{D}$ denotes the batch of transitions following $\pi_{p}$. 
By introducing a vector of additional variables $\boldsymbol{z_{p}} = (z_{1,p}^2, z_{2,p}^2,\cdots,z_{m,p}^2, z_{m+1,p}^2)$, we can convert the Problem (\ref{RL-based constrained optimization with inequality constraints}) to the following problem:
\begin{equation}\label{RL-based constrained optimization with equality constraints b adding additional variables}
        \begin{split}
            \min_{\theta_p, \boldsymbol{z_p}} \,\, & -V^{\theta_p}\\
            s.t. 
            \,\,& \mathbb{E}_{x_{t_k}\sim \mu_{\pi_{p}},u_{t_k}\sim \pi_p} \! \big[-\Phi_{i,r}(x_{t_k})\big] + z_{i,p}^2 \! =  \! 0 \,\,\,\, \forall i \in [1,m],\vspace{1ex} \\\vspace{1ex}
            \,\,& \mathbb{E}_{\!x_{t_k}\!\sim \mu_{\pi_{p}},u_{t_k}\!\sim \pi_p} \! \big[\! L_{\!\nu\!}(\!\hat{x}_{t_{k+1}}\!)  \!\! -\!  \! L_{\!\nu\!}(\!x_{t_k}\!)  \!\! + \! \! \beta L_{\!\nu\!}(\!x_{t_k}\!)\big] \!+ \!z_{\!m+1\!,p\!}^2 \! =  \! 0.
        \end{split}
    \end{equation}
Denote the Lagrangian multipliers for CBF and CLF constraints for this primary controller as $\lambda_{i,p}$ and $\zeta$, respectively, and the penalty parameter as $c_p$. Furthermore, define $f_{i,p}(\theta_p) \triangleq \mathbb{E}_{x_{t_k}\sim \mu_{\pi_p},u_{t_k}\sim \pi_p} \! \big[  -\Phi_{i,r}(x_{t_k})\big]$, $\forall i \in [1,m]$, and $g(\theta_p) \triangleq \mathbb{E}_{x_{t_k}\sim \mu_{\pi_p},u_{t_k}\sim \pi_p} \! \big[L_{\nu}(\hat{x}_{t_{k+1}})  \! -  \! L_{\nu}(x_{t_k})  \! +  \! \beta L_{\nu}(x_{t_k})\big]$, the augmented Lagrangian function is then $\mathcal{L}^{p}_{c_p} (\theta_p,\lambda_{i,p},\zeta) = -V^{\theta_p}+\!\sum_{i = 1}^{m} \lambda_{i,p}\! \times\! \large(f_{i,p}(\theta_p) + z_{i,p}^2 \large)+ \!\sum_{i = 1}^{m}\frac{c_p}{2}\! \times\!\large(f_{i,p}(\theta_p) + z_{i,p}^2 \large)^2+\!\zeta\! \!\times \!\large(g(\theta_p) + z_{m+1,p}^2 \large) +\! \frac{c_p}{2} \! \times \! \large(g(\theta_p) + z_{m+1,p}^2 \large)^2$. The updates of parameters are in Algorithm \ref{alg:NLBAC}.\\
\begin{algorithm}
    \caption{Neural Ordinary Differential Equations-based Lyapunov-Barrier Actor-Critic}
    \label{alg:NLBAC}
    \begin{algorithmic}[2]

    \renewcommand{\algorithmicrequire}{\textbf{Input:}}
    \renewcommand{\algorithmicensure}{\textbf{Output:}}
    \REQUIRE Number of steps $N=0$, an initialized NODE model parameterized by $\varphi$, action-value networks $Q^{\pi_{p}}_{\phi_i}$, Lyapunov network $L_{\nu}$, primary controller network $\pi_{\theta_p}$, coefficient $\alpha_p$ and Lagrange multipliers $\lambda_{i,p}$ and $\zeta$ for the primary controller, RL-based backup controller network $\pi_{\theta_b}$, coefficient $\alpha_b$ and Lagrange multipliers $\lambda_{i,b}$ for the backup controller, replay buffer $\mathcal{B}$, coefficients of quadratic terms $c_p$ and $c_b$, quadratic term coefficient factor $\rho_c \in (1,\infty)$, learning rates $\eta_{1}$, $\eta_{2}$, $\eta_{3}$. $n_m$, $n_L$, and $n_b$ which are delay steps for NODE model, Lagrangian multipliers, and backup controller updates, respectively.

    \FOR{each episode}
        \FOR{each step}
            \STATE $N \leftarrow N + 1$
            \IF{$N\, \mathrm{mod}\, n_m = 0$}
                \STATE Update the NODEs model with data collected during the controller learning process:
                \STATE $\psi \leftarrow \psi - \eta_{1} \nabla_{\psi} \ell $
            \ENDIF
            \STATE Construct CBF and CLF constraints with the NODE model and transition pairs from $\mathcal{B}$
            \STATE Update the Lyapunov network and action-value networks:
            \STATE $\nu\leftarrow \nu - \eta_{2}\nabla_{\nu}J_{L}(L_{\nu})$; $\phi_{i}\leftarrow \phi_{i} - \eta_{2}\nabla_{\phi_{i}}J_{Q^{\pi_{p}}}(Q^{\pi_{p}}_{\phi_{i}})$ for $i \in \{1,2\}$
            \STATE Update the primary controller network and its coefficient $\alpha_p$:
            \STATE $\theta_p \leftarrow \theta_p - \eta_{3}\nabla_{\theta_p}\mathcal{L}^{p}_{c_p} (\theta_p,\lambda_{i,p},\zeta)$; $\alpha_p \leftarrow \alpha_{p} - \eta_{3}\nabla_{\alpha_p}J_{\alpha_p}(\alpha_p)$
            \STATE $c_p \leftarrow \rho_c \times c_p$   
            \IF{$N\, \mathrm{mod}\, n_L = 0$}
                \STATE Update the Lagrangian multipliers $\lambda_{i,p}$ and $\zeta$ according to Bertsekas et al. \cite{bertsekas1982constrained}:
                \STATE $\lambda_{i,p} \leftarrow \mathrm{max} \{ 0, \lambda_{i,p} + c_p f_{i,p}(\theta_p) \}  $; \\
                $\zeta \leftarrow \mathrm{max} \{ 0, \zeta + c_p g(\theta_p) \}  $
            \ENDIF
            \IF{$N\, \mathrm{mod}\, n_b = 0$}
                \STATE Update the backup controller network and its coefficient $\alpha_b$:
                \STATE $\theta_b \leftarrow \theta_b - \eta_{3}\nabla_{\theta_b}\mathcal{L}^{b}_{c_b} (\theta_b,\lambda_{i,b})$; \\
                $\alpha_b \leftarrow \alpha_{b} - \eta_{3}\nabla_{\alpha_b}J_{\alpha_b}(\alpha_b)$
                \STATE $c_b \leftarrow \rho_c \times c_b$ 
                \IF{$N\, \mathrm{mod}\, (n_b \times n_L) = 0$}
                    \STATE Update the Lagrangian multipliers $\lambda_{i,b}$:
                    \STATE $\lambda_{i,b} \leftarrow \mathrm{max} \{ 0, \lambda_{i,b} + c_b f_{i,b}(\theta_b) \}  $
                \ENDIF
            \ENDIF
            \IF{Backup controller should be used according to the condition specific to the task}
                \STATE $u_{t_k} \sim \pi_{\theta_b}(u_{t_k} | x_{t_k})$ and apply control signal $u_{t_k}$
            \ELSE
                \STATE $u_{t_k} \sim \pi_{\theta_p}(u_{t_k} | x_{t_k})$ and apply control signal $u_{t_k}$
                \STATE Store the transition pair $(x_{t_k},u_{t_k},r_{t_k},c_{t_k},x_{t_{k+1}})$ in $\mathcal{B}$
            \ENDIF
        \ENDFOR
    \ENDFOR
    \ENSURE{$\pi_{\theta_p}$, $\pi_{\theta_b}$, $Q^{\pi_{p}}_{\phi_i}, i=1,2$, and $L_{\nu}$}
    \end{algorithmic}
\end{algorithm}
We also introduce some tricks used in implementations. Firstly, a different timescale method is applied similar to Yu et al. \cite{yu2022reachability}, and furthermore, we set $n_m$, $n_L$, and $n_b$ to update the NODEs model, Lagrangian multipliers and the backup controller with delayed update tricks \cite{ma2021feasible}. Also, in practical implementation, we sample transition pairs from the replay buffer $\mathcal{B}$ for updating parameters, and before calculating the expected values for the constraints in Problem (\ref{RL-based constrained optimization with inequality constraints}), we first apply the ReLU function to CBF and CLF constraints at each sampled $x_{t_k}$.

\subsection{Backup Controller Design}
Due to the existence of multiple constraints, the feasibility of the Problem (\ref{RL-based constrained optimization with inequality constraints}) becomes a crucial problem. Priority is given to safety constraints when both constraints cannot be satisfied simultaneously, therefore, we design an RL-based backup controller $\pi_b$ parameterized by $\theta_b$ by formulating an additional constrained optimization problem as follows:
\begin{equation}\label{backup RL-based constrained optimization with inequality constraints}
        \begin{split}
            \min_{\theta_b} \,\, & -V^{\theta_b}\\
            s.t. 
            \,\,& \mathbb{E}_{x_{t_k}\sim \mu_{\pi_p},u_{t_k}\sim \pi_b} \! \big[-\Phi_{i,r}(x_{t_k})\big]  \! \leq  \! 0 \,\,\, \forall i \in [1,m],\vspace{1ex} 
        \end{split}
    \end{equation}
and the predicted states are now following the backup controller $\pi_{b}$. The objective function used for the backup controller is used to maximize the cumulative discounted reward if we use the backup controller $\pi_b$ for one step. \\
 For simplicity, we define $f_{i,b}(\theta_b) \triangleq \mathbb{E}_{x_{t_k}\sim \mu_{\pi_p},u_{t_k}\sim \pi_b} \! \big[-\Phi_{i,r}(x_{t_k})\big]$, $\forall i \in [1,m]$. Therefore, the augmented Lagrangian function for updating the backup controller is given as $\mathcal{L}^{b}_{c_b} (\theta_b,\lambda_{i,b}) = -V^{\theta_b}+\!\sum_{i = 1}^{m} \lambda_{i,b}\! \times\! \large(f_{i,b}(\theta_b) + z_{i,b}^2 \large)+ \!\sum_{i = 1}^{m}\frac{c_b}{2}\! \times\!\large(f_{i,b}(\theta_b) + z_{i,b}^2 \large)^2$, where $\lambda_{i,b}$, $c_b$ and $\boldsymbol{z_{b}} = (z_{1,b}^2, z_{2,b}^2,\cdots,z_{m,b}^2)$ are Lagrangian multipliers for CBF constraints, the penalty parameter, and additional variables. Also, in real implementation, we sample the data from the replay buffer $\mathcal{B}$, and apply the ReLU function to CBF constraints at each sampled $x_{t_k}$. The framework combining the primary and backup controllers can be summarized as Algorithm \ref{alg:NLBAC}.

\section{Simulations} 
\label{sec:experiment}
\begin{figure}[tb]
        \centering
        \includegraphics[scale=0.0816]{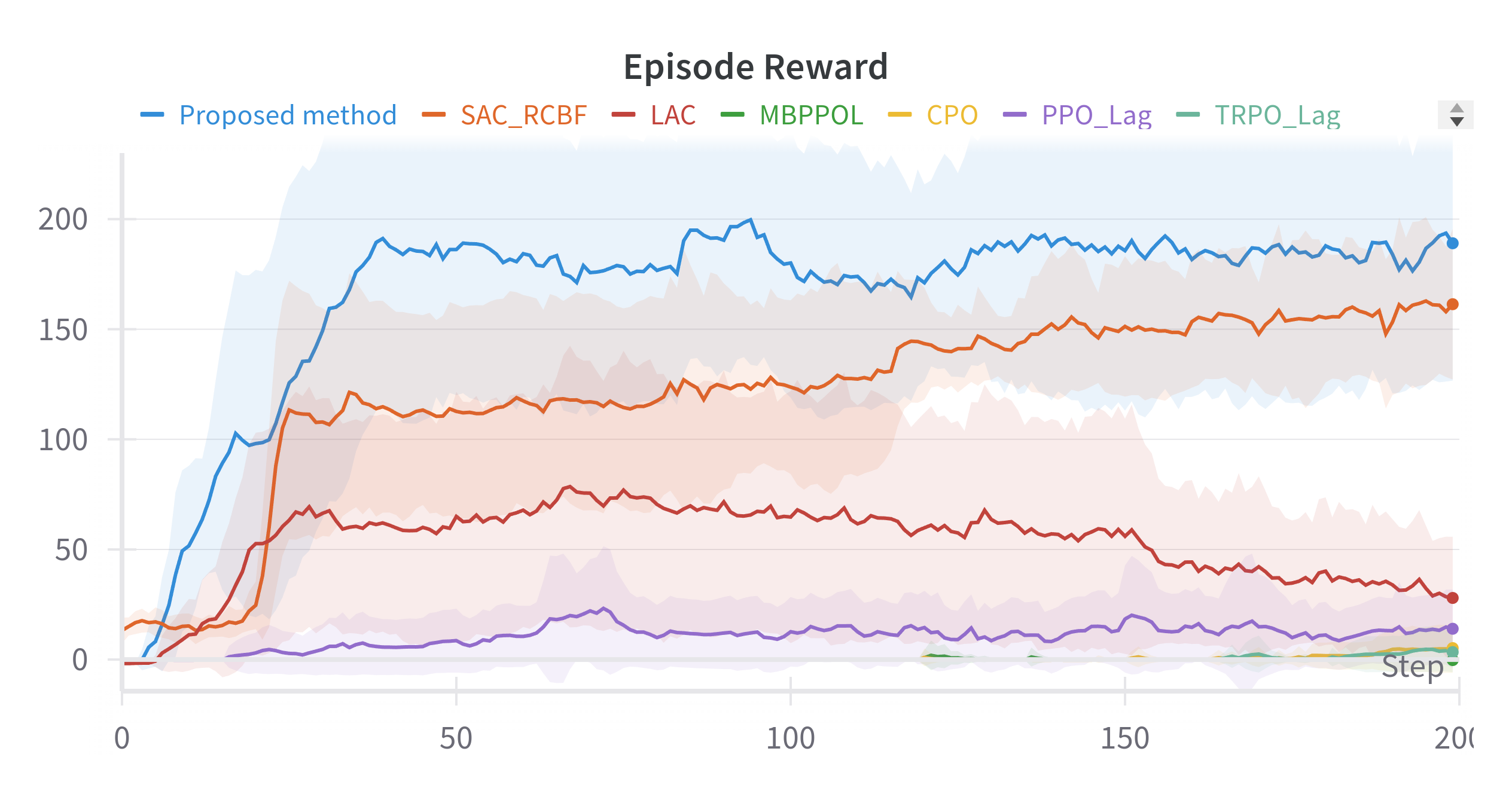}
        \hspace{5mm}
        \includegraphics[scale=0.0816]{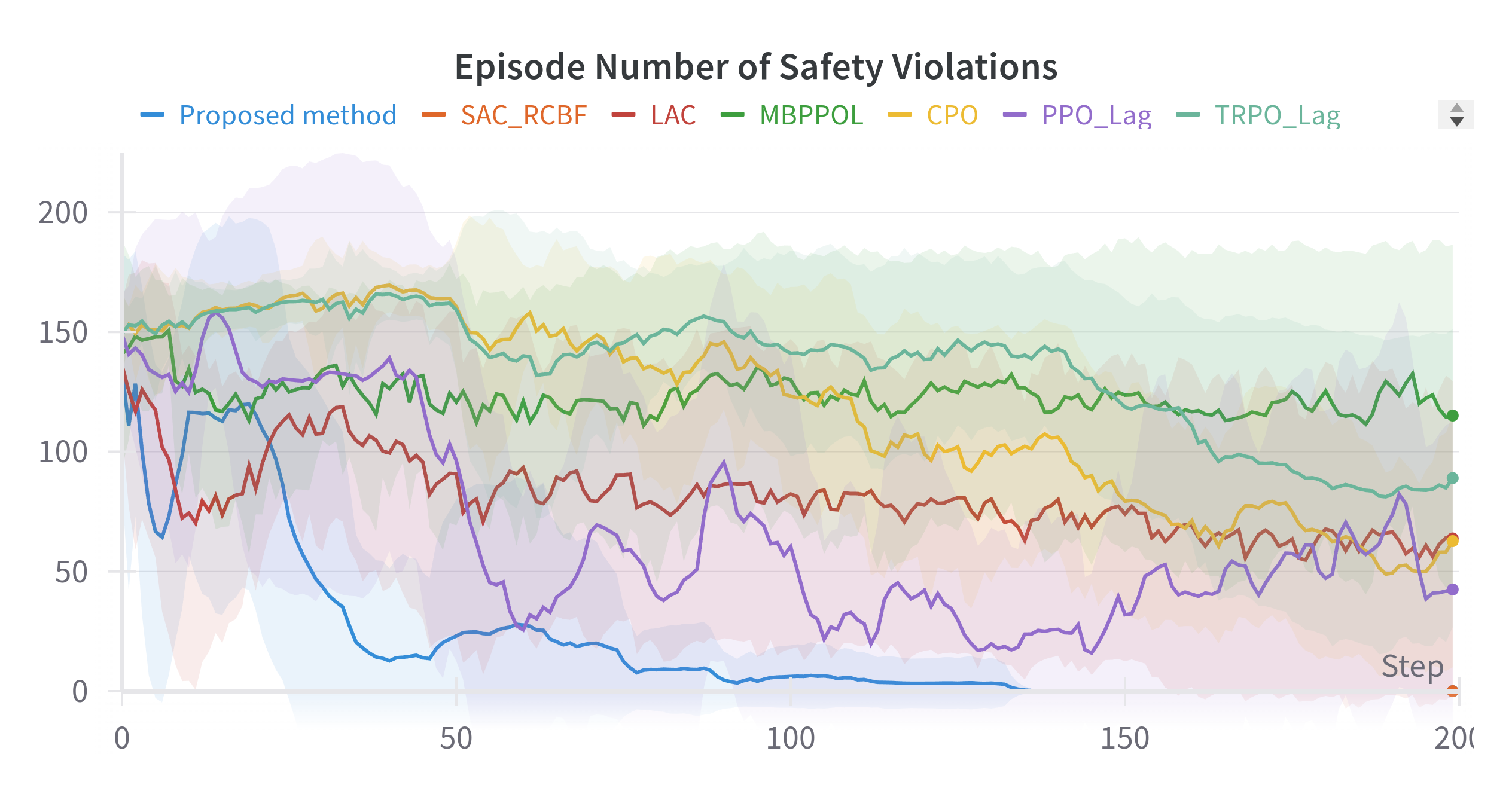}
        \caption{The cumulative reward and cumulative number of safety violations of each episode in the simulated car following environment setting are compared between the proposed method (drawn in blue) and baselines. Each curve illustrates the average across ten experiments employing different random seeds, with the shaded area denoting the standard deviation. The safety violation of the SAC-RCBF algorithm keeps being 0 and therefore its graph is coincident with the X-axis.}
        \label{fig:Simulated Car Following results}
\vspace{-0.7cm}     
\end{figure}

Currently, there are many studies applying machine learning algorithms to solve problems in the field of transportation \cite{ma2022traffic, wang2022work,ma2023ramp,miao2023volumetric,wang2023transfer,su2023uncertainty,lin2024drplanner}, and therefore, we conduct experiments on the task called ``Simulated Car Following", and we use SAC-RCBF \cite{emam2021safe}, MBPPO-Lagrangian \cite{jayant2022model}, LAC \cite{han2020actor}, CPO \cite{achiam2017constrained}, PPO-Lagrangian and TRPO-Lagrangian \cite{Ray2019} as baselines. The environment ``Simulated Car Following" is modified (but different) from Zhao et al. \cite{zhao2023barrier}. This task involves a chain of five cars, four of which are driven by human drivers while the $4^{th}$ one is the robot car controlled by an RL-based controller. These five cars are following each other on a straight road, and the goal is to control the acceleration of the $4^{th}$ car to maintain a desired distance from the $3^{rd}$ car while avoiding collisions with other cars, namely achieving the goal state while maintaining safety when human drivers exist. The movements of cars driven by human drivers are given by:
\begin{equation*}
    \begin{array}{l}
        \dot{x}_{t_{k},i}=
        \left[ \begin{array}{c}
        v_{t_{k},i} \\
        0
        \end{array}
        \right ] +
        \left[ \begin{array}{c}
            0  \\
            1+d_i 
            \end{array}
            \right ] a_{t_{k},i}   \qquad  \forall i \in \{1,2,3,5\}.
    \end{array}
\end{equation*}
Each state of the system is denoted as $x_{t_{k},i} = [p_{t_{k},i},v_{t_{k},i}]^T$, indicating the position $p_{t_{k},i}$ and velocity $v_{t_{k},i}$ of the $i^{th}$ car at the timestep $t_{k}$, $d_i=0.1$. The time interval used in this experiment is $0.02$s. The predefined velocity of the $1^{st}$ car is $v_{s} - 4\sin (t_{k})$, where $v_{s}=3.0$. Its acceleration is given as $a_{t_{k},1}=k_v(v_{s} - 4\sin (t_{k}) -v_{t_{k},1})$ where $k_v=4.0$. The human drivers of the following cars decide the accelerations of the $2^{nd}$, $3^{rd}$, and $5^{th}$ car, and can change their decisions sharply.  Accelerations of Car 2 and 3 are given by:
    \begin{equation*}
    a_{t_{k},i}\!=\!\left\{
\begin{aligned}  
&\!k_v(v_{s}\!-\!v_{t_{k},i})\!-\!k_b(p_{t_{k},i\!-\!1}\!-\!p_{t_{k},i})\,\,if\,|p_{t_{k},i\!-\!1}\!-\!p_{t_{k},i}|\! <\! 6.5\\
&\!k_v(v_{s}\!-\!v_{t_{k},i})\,\,\,\,\,\,\,\,\,\,\,\,\,\,\,\,\,\,\,\,\,\,\,\,\,\,\,\,\,\,\,\,\,\,\,\,\,\,\,\,\,\,\,\,\,otherwise, \\
\end{aligned}
\right.
\end{equation*}    
where $k_b=20.0$ and $i=2,3$. The acceleration of the $5^{th}$ car is:
\begin{equation*}
    a_{t_{k},5}\!=\!\left\{
\begin{aligned}
&\!k_v(v_{s}\!-\!v_{t_{k},5})\!-\!k_b(p_{t_{k},3}\!-\!p_{t_{k},5})\,\,if\,|p_{t_{k},3}-p_{t_{k},5}| \!<\! 13.0\\
&\!k_v(v_{s}\!-\!v_{t_{k},5})\,\,\,\,\,\,\,\,\,\,\,\,\,\,\,\,\,\,\,\,\,\,\,\,\,\,\,\,\,\,\,\,\,\,\,\,\,\,\,\,\,\,\,\,\,otherwise, \\
\end{aligned}
\right.
\end{equation*}
The model of the $4^{th}$ car is as follows: 

        \begin{equation*}
         \dot{x}_{t_k,4}=
         \left[ \begin{array}{c}
         v_{t_{k},4} \\
         0
         \end{array}
         \right ] +
         \left[ \begin{array}{c}
         0 \\
         1.0
         \end{array}
         \right ] u_{t_{k}} ,
         \end{equation*} 
where $u_{t_{k}}$ is the acceleration of the $4^{th}$ car (robot), and also the control signal generated by the RL-based controller at the timestep $t_{k}$. The reward signal is defined to minimize the overall control effort, and an additional reward of 2.0 is granted during timesteps when $d_{t_k} = p_{t_k,3}-p_{t_k,4}$, which is the distance between the $3^{rd}$ and $4^{th}$ car, falls within $[9.0,10.0]$. This range is defined as the desired region for $d_{t_k}$, and as this task rewards the system when it tries to maintain the desired state (distance), we employ cumulative reward as a metric, and a higher cumulative reward indicates better convergence to the desired goal state. The cost signal is determined as $\left\lVert d_{t_{k+1}} - d_{\text{desired}}\right\rVert $, where $d_{\text{desired}}=9.5$. CBFs are defined as $h_1(x_{t_k}) =p_{t_{k},3}-p_{t_{k},4} - \delta$ and $h_2(x_{t_k}) =p_{t_{k},4}-p_{t_{k},5} - \delta$, with $\delta$ being the minimum required distance between the cars. Hence, the relative degree is 2 and the planning horizon for making predictions using NODEs is 2. When a safety constraint is violated if two types of constraints cannot be satisfied simultaneously, the $4^{th}$ car might be in close proximity to the $5^{th}$ human driver in order to make $d_{t_k}$ be within $[9.0,10.0]$. In such cases, the backup controller is activated. The primary controller is reactivated when the $4^{th}$ car moves beyond the dangerous area, namely out of the vicinity of the $5^{th}$ car, or when the predetermined time threshold for utilizing the backup controller is exceeded.\\
Noted that when we use NODEs to model this system, the input of the network $\mathcal{F}$ which is structured by MLP is $\left(t_k, \hat x_{t_k}, u_{t_k}\right)$ with the dimension of 12, and the output dimension is 10. As illustrated in Fig. \ref{fig:Simulated Car Following results}, the proposed method consistently achieves the highest cumulative reward. This outcome indicates its exceptional capability in effectively regulating the distance $d_{t_k}$ within the desired range $[9.0,10.0]$, which is the desired goal. Moreover, the cumulative number of safety violations caused by the proposed method is considerably smaller than those of others after some episodes and finally decreases to almost 0. Even compared to the SAC-RCBF algorithm where a good nominal model is used and Gaussian processes (GPs) are applied to approximate the differences between the real and nominal dynamics, our method which does not require prior knowledge (for example, a nominal model) of the system dynamics always achieves higher reward during the whole training, and comparatively good performance in maintaining zero safety violations in the latter part of the training process. This shows that our method can achieve and maintain both safety and stability in fewer iterations of training in a human-robot scenario compared to the model-based and model-free baselines, which makes this proposed method appropriate to be utilized in real-world applications where algorithms should be sample efficient. It is also noteworthy that methods using GPs to help approximate the dynamics \cite{zhao2023barrier} face the potential problem of having a larger computational burden since GP is a non-parametric method that scales poorly with the size of the data collected, while NODE does not have this problem.
\section{CONCLUSIONS}
\label{sec:conclusion}
This paper introduces a method that helps guarantee both the safety and stability for human-robot scenarios where the movements of both human and robot are approximated by NODEs, and the experiments show higher cumulative rewards and fewer safety violations with better sample efficiency.\\
However, this method has limitations. For example, although good performance is achieved in our task, the difference between the real dynamics and NODE-based models is still unknown and can be large in human-aligned tasks which are more complex, and this may harm the performance of the proposed method. We believe that addressing current limitations could be
interesting future directions.

\bibliographystyle{IEEEtran}
\bibliography{bibli_new}

\end{document}